\documentclass{article}

\PassOptionsToPackage{compress, round}{natbib}

\usepackage[final]{nips_2017}

\usepackage[utf8]{inputenc}       % allow utf-8 input
\usepackage[T1]{fontenc}          % use 8-bit T1 fonts
\usepackage[hidelinks]{hyperref}  % hyperlinks
\usepackage{url}                  % simple URL typesetting
\usepackage{booktabs}             % professional-quality tables
\usepackage{amsfonts}             % blackboard math symbols
\usepackage{nicefrac}             % compact symbols for 1/2, etc.
\usepackage{microtype}            % microtypography

\usepackage[nolist]{acronym}
\usepackage{amsmath}
\usepackage[english]{babel}
\usepackage[labelfont=bf]{caption}  % for bold caption titles
\usepackage[autostyle]{csquotes}    % for the quotes: >>...<<
\usepackage{enumitem}
\usepackage{ifthen}
\usepackage{multirow}
\usepackage{pgfplots}
\newcommand{\mat}[1]{\mathbf{#1}}   % format for matrices
\newcommand{\R}{\mathbb{R}}         % the symbol for the reals
\newcommand{\vect}[1]{\mathbf{#1}}  % format for vectors
\newcommand{\MyQuote}[1]{\frqq\textit{#1}\flqq}

\newcommand{\MyUrl}[1]{\small{\url{#1}}}

\newcommand{\MyAcf}[2][]
{%
    \aclu{#2} (\acs{#2}\ifthenelse{\equal{#1}{}}{}{; #1})%
}

\title{Deep Learning for Ontology Reasoning}

\author{
    Patrick Hohenecker, Thomas Lukasiewicz\\
    Department of Computer Science\\
    University of Oxford\\
    Oxford, United Kingdom\\
    \small{%
        \{%
            \href{mailto:patrick.hohenecker@cs.ox.ac.uk}{\tt{patrick.hohenecker}},
            \href{mailto:thomas.lukasiewicz@cs.ox.ac.uk}{\tt{thomas.lukasiewicz}}%
        \}%
        \texttt{@cs.ox.ac.uk}
    }\\
}

\begin{document}

\maketitle

% //////// ABSTRACT //////////////////////////////////////////////////////////////////////////////////

\begin{abstract}
    In this work, we present a novel approach to ontology reasoning that is based on deep learning
    rather than logic-based formal reasoning.
    To this end, we introduce a new model for statistical relational learning that is built upon deep
    recursive neural networks, and give experimental evidence that it can easily compete with, or even
    outperform, existing logic-based reasoners on the task of ontology reasoning.
    More precisely, we compared our implemented system with one of the best logic-based ontology
    reasoners at present, \mbox{RDFox}, on a number of large standard benchmark datasets, and found
    that our system attained high reasoning quality, while being up to two orders of magnitude
    faster.
\end{abstract}

% //////// INTRODUCTION //////////////////////////////////////////////////////////////////////////////

\section{Introduction}\label{sec:introduction}

In the last few years, there has been an increasing interest in the application of \ac{ML} to the
field of \ac{KRR}, or, more generally, in learning to reason over symbolic data---cf., e.g.,
\citet{Gabrilovoch2015}.
The main motivation behind this is that most \ac{KRR} formalisms used today are rooted in symbolic
logic, which allows for answering queries accurately by employing formal reasoning, but also comes
with a number of issues, like difficulties with handling incomplete, conflicting, or uncertain
information and scalability problems.

However, many of these issues can be dealt with effectively by using methods of \ac{ML}, which are in
this context often subsumed under the notion of
\MyAcf[\citealp{Getoor2007}]{SRL}---cf.\ \citet{Nickel2016} for a recent survey.
Notice, though, that the use of \ac{ML} for reasoning is a tradeoff.
On the one hand, ML models are often highly scalable, more resistant to disturbances in the data, and
can provide predictions even if formal reasoning fails.
On the other hand, however, their predictions are correct with a certain probability only.
In contrast to this, formal reasoners are often obstructed by the above problems, but if they can
provide inferences, then these are correct with certainty.

We believe that the combination of both fields, i.e., \ac{ML} and \ac{KRR}, is an important step
towards human-level artificial intelligence.
However, while there exist elaborate reasoning systems already, \ac{SRL} is a rather young field that
has, we believe, not hit its boundaries yet.
Therefore, in this work, we introduce a new approach to \ac{SRL} based on deep learning, and apply it
to the task of reasoning over \acp{OKB}.
These are \acp{KB} that consist of a set of facts together with a formal description of the domain of
interest---the so-called ontology.
The reason why we chose this very task is its practical significance as well as the fact that it
commonly comprises extensive formal reasoning.

The motivation for employing deep learning, however, which refers to the use of \acp{NN} that perform
many sequential steps of computation, should be fairly obvious.
In the last ten years, deep learning has been applied to a wide variety of problems with tremendous
success, and constitutes the state-of-the-art in fields like computer vision and \ac{NLP} today.
Interestingly, there are also a few published attempts to realize formal reasoning by means of deep
\acp{NN}.
However, these focus on rather restricted logics, like natural logic \citep{Bowman2013} or real logic
\citep{Serafini2016}, and do not consider reasoning in its full generality.
Besides this, \MyQuote{reasoning} appears in connection with deep learning mostly in the context of
\ac{NLP}---e.g., \citet{Socher2013}.

%In practice, and in the context of description logics \citep{Baader2007}, ontologies are usually
%defined in terms of unary and binary predicates only.
%Therefore, we also confine ourselves to this particular case here---our approach can be easily
%extended to the general case of predicates of higher arity, though.

The main contributions of this paper are briefly as follows: 
\begin{itemize}[parsep=0pt, itemsep=2pt]
    \item We present a novel method for \ac{SRL} that is based on deep learning with recursive
          \acp{NN}, and apply it to ontology reasoning. 
    \item Furthermore, we provide an experimental comparison of the suggested approach with one of the
          best logic-based ontology reasoners at present, \mbox{RDFox} \citep{Nenov2015}, on several
          large standard benchmarks.
          Thereby, our model achieves a high reasoning quality while being up to two orders of
          magnitude faster.
    \item To the best of our knowledge, we are the first to investigate ontology reasoning based on
          deep learning on such large and expressive \acp{OKB}.
\end{itemize}

The rest of this paper is organized as follows.
In the next section, we review a few concepts that our approach is built upon.
Section \ref{sec:rtn} introduces the suggested model in full detail, and Section \ref{sec:reasoning}
discusses how to apply it to ontology reasoning.
In Section~\ref{sec:evaluation}, we evaluate our model on four datasets, and compare its performance
with RDFox.
We conclude with a summary of the main results, and give an outlook on future research.

%\textcolor{red}{Both sides benefit (ML and KR\&R).}

% //////// BACKGROUND ////////////////////////////////////////////////////////////////////////////////

\section{Background}\label{sec:background}

As mentioned in the introduction already, our work lies at the intersection of two, traditionally
quite separated, fields, namely \ac{ML} and \ac{KRR}.
Therefore, in this section, we review the most important concepts, from both areas, that are required
to follow the subsequent elaborations.

% //////// ONTOLOGICAL KNOWLEDGE BASES ---------------------------------------------------------------

\subsection{Ontological Knowledge Bases (OKBs)}\label{sec:okbs}
\acused{OKB}

A central idea in the field of \ac{KRR} is the use of so-called ontologies.
In this context, an ontology is a formal description of a concept or a domain, e.g., a part of the
real world, and the word \MyQuote{formal} emphasizes that such a description needs to be specified by
means of some knowledge representation language with clearly defined semantics.
This, in turn, allows us to employ formal reasoning in order to draw conclusions based on such an
ontology.

An important aspect to note is that an ontology is situated on the meta-level, which means that it
might specify general concepts or relations, but does not contain any facts.
However, in the sequel we only talk about a number of facts together with an ontology that describes the domain of interest, and we refer to such a setting as an \acf{OKB}.

In practice, and in the context of description logics \citep{Baader2007}, ontologies are usually
defined in terms of unary and binary predicates.
Thereby, unary predicates are usually referred to as concepts or classes, and define certain
categories, e.g., of individuals that possess a particular characteristic.
In contrast to this, binary predicates define relationships that might exist between a pair of
individuals, and are usually referred to as relations or roles.

What is really appealing about ontologies is that they usually not just define those predicates, but
also rules that allow us to draw conclusions based on them.
This could encompass simple inferences like every individual of class \texttt{women} belongs to class
\texttt{human} as well, but also much more elaborate reasoning that takes several classes and
relations into account.
Notice further that we can view almost any relational dataset as an \ac{OKB} with an ontology that
does not specify anything except the classes and relations that exist in the data.

Based on the fact that we hardly ever encounter ontologies with predicates of arity greater than two
in practice, we confine ourselves to this particular case in the subsequent treatment---the approach
introduced in this work can be easily extended to the general case, though.
Any \ac{OKB} that is defined in terms of unary and binary predicates only has a natural
representation as labeled directed multigraph\footnote{%
    If we really need to account for predicates of arity greater than two, then we can view any such
    dataset as a hypergraph, and extend the \acs{RTN} model introduced in the next section with
    convolutional layers as appropriate.
}
if individuals are interpreted as vertices and every occurrence of a binary predicate as a directed
edge.
Thereby, edges are labeled with the name of the according relation, and vertices with an incidence
vector that indicates which classes they belong to.
Notice, however, that, depending on the used formalism, \acp{OKB} may adhere to the so-called
\ac{OWA}.
In this case, a fact can be \textbf{true}, \textbf{false}, or \textbf{unknown}, which is, e.g.,
different from classical first-order logic.
The presence of the \ac{OWA} is reflected by according three-valued incidence vectors, whose elements
may be any of $1$, $-1$, or $0$, respectively, and indicate that an individual belongs to a class, is
not a member of the same, or that this is unknown.

% //////// RECURSIVE NEURAL TENSOR NETWORKS ----------------------------------------------------------

\subsection{Recursive Neural Tensor Networks (RNTNs)}\label{sec:rntns}
\acused{RNTN}

Recursive \acp{NN} \citep{Pollack1990} are a special kind of network architecture that was introduced
in order to deal with training instances that are given as trees rather than, as more commonly,
feature vectors.
In general, they can deal with any \ac{DAG}, since any such graph can be unrolled as a tree, and the
only requirement is that the leaf nodes have vector representations attached to them.
An example from the field of \ac{NLP} is the parse tree of a sentence, where each node represents one
word and is given as either a one-hot-vector or a previously learned word embedding.

Unlike feed-forward networks, recursive \acp{NN} do not have a fixed network structure, but only
define a single recursive layer, which accepts two vectors as input and maps them to a common
embedding.
This layer is used to reduce a provided tree step by step in a bottom-up fashion until only one
single vector is left.
The resulting vector can be regarded as an embedding of the entire graph, and may be used, e.g., as
input for a subsequent prediction task.

In this work, we make use of the following recursive layer, which defines what is referred to as
\MyAcf[\citealp{Socher2013}]{RNTN}:
\begin{equation}\label{eq:original-tensor-layer}
	g ( \vect{x} , R , \vect{y} ) =
    \mat{U}_R f \left(
        \vect{x}^T \mat{W}_R^{[ 1 : k ]} \vect{y} +
        \mat{V}_R \begin{bmatrix} \vect{x} \\ \vect{y} \end{bmatrix} +
        \vect{b}_R
    \right)
    \text{,}
\end{equation}
where
$\vect{x} , \vect{y} \,{\in}\, \R^d$\!,
$\mat{U}_R \,{\in}\, \R^{d \times k}$\!,
$\mat{V}_R \,{\in}\, \R^{k \times 2 d}$\!,
$\mat{W}_R  \,{\in}\, \R^{d \times d \times k}$\!,
$\mat{b}_R  \,{\in}\, \R^k$\!, and
$f$ is a nonlinearity that is applied element-wise, commonly $\tanh$.
Thereby, the term $\vect{x}^T \mat{W}_R^{[ 1 : k ]} \vect{y}$ denotes a bilinear tensor product, and
is computed by multiplying $\vect{x}$ and~$\vect{y}$ with every slice of $\mat{W}_R$ separately.
So, if $\vect{z}$ is the computed tensor product, then
$\vect{z}_i = \vect{x}^T \mat{W}_R^{[i]} \vect{y}$.
In addition to the actual input vectors, $\vect{x}$ and $\vect{y}$, the tensor layer accepts another
parameter $R$, which may be used to specify a certain relation between the provided
vectors.
This makes the model more powerful, since we use a separate set of weights for each kind of relation.

In general, recursive \acp{NN} are trained by means of \ac{SGD} together with a straightforward extension of standard backpropagation, called \MyAcf[\citealp{Goller1996}]{BPTS}.

% //////// RELATIONAL TENSOR NETWORKS ////////////////////////////////////////////////////////////////

\section{Relational Tensor Networks (RTNs)}\label{sec:rtn}
\acused{RTN}

In this section, we present a new model for \ac{SRL}, which we---due to lack of a better name---refer
to as \MyAcf{RTN}.
An \ac{RTN} is basically an \ac{RNTN} that makes use of a modified bilinear tensor layer.
The underlying intuition, however, is quite different, and the term \MyQuote{relational}
emphasizes the focus on relational datasets.

% //////// THE BASIC MODEL ---------------------------------------------------------------------------

\subsection{The Basic Model}

As described in the previous section, recursive \acp{NN} allow for computing embeddings of training
instances that are given as \acp{DAG}.
If we face a relational dataset, though, then the training samples are actually vertices of a graph,
namely the one that is induced by the entire relational dataset, rather than a graph itself.
However, while this does not fit the original framework of recursive networks, we can still make use
of a recursive layer in order to update the representations of individuals based on the structure of
dataset.
In an \ac{RTN}, this deliberation is reflected by the following modified tensor layer:
\begin{equation}\label{eq:tensor-layer}
	\tilde{g} ( \vect{x} , R , \vect{y} ) =
    \vect{x} +
    \mat{U}_R f \left(
        \vect{x}^T \mat{W}_R^{[ 1 : m ]} \vect{y} +
        \mat{V}_R \vect{y}
    \right)
    \text{,}
\end{equation}
where the notation is the same as in Equation \ref{eq:original-tensor-layer} except that
$\mat{V}_R \,{\in}\, \R^{k \times d}$.

The intuition here is quite straightforward. While individuals in a relational dataset are initially represented by their respective
feature vectors, big parts of the total information that we have are actually hidden in the relations among them.
However, we can use a recursive network, composed of tensor layers like the one denoted in Equation
\ref{eq:tensor-layer}, to incorporate these data into an individual's embedding.
Intuitively, this means that we basically apply a recursive \ac{NN} to an update tree of an
individual, and thus compute an according vector representation based on the relations that it is
involved in.
%The resulting vectors can be considered as learned embeddings of the individuals in the dataset, and
%may be used as input for subsequent prediction tasks.
For the \ac{RTN}, we adopted the convention that a tensor layer
$\tilde{g}$ updates the individual represented by $\vect{x}$ based on an instance
$( \vect{x} , R , \vect{y} )$ of relation $R$ that is present in the data.
Furthermore, if the relations in the considered dataset are not symmetric, then we have to distinguish
whether an individual is the source or the target of an instance of a relation.
Accordingly, the model has to contain two sets of parameters for such a relation, one for updating
the source and one for the target, and we denote these as $R^\triangleright$ and $R^\triangleleft$,
respectively.
This means, e.g., that $\tilde{g} ( \vect{x} , R^\triangleleft , \vect{y} )$ denotes that the
embedding of $\vect{x}$ is updated based on $( \vect{y} , R , \vect{x} )$.

The foregoing considerations also explain the differences between Equation \ref{eq:tensor-layer} and
the original tensor layer given in Equation \ref{eq:original-tensor-layer} \citep{Socher2013}.
First and foremost, we see that in our model $\vect{x}$ is added to what basically used to be the
tensor layer before, which is predicated on the fact that we want to update this very vector.
Furthermore, $\vect{x}$ does not affect the argument of the nonlinearity $f$ independently of
$\vect{y}$, since $\vect{x}$ by itself should not determine the way that it is updated.
Lastly, there is no bias term on the right-hand side of Equation \ref{eq:tensor-layer} to prevent
that there is some kind of default update irrespective of the individuals involved.

We also considered to add another application of the hyperbolic tangent on top of the calculations
given in Equation \ref{eq:tensor-layer} in order to keep the elements of the created embeddings in
$[ -1 ,1 ]$.
This would ensure that there cannot be any embeddings with an oddly large norm due to individuals
being involved in a large number of relations.
However, since we did not encounter any problems like this in our experiments, we decided against the
use of this option, as it could introduce additional problems like vanishing gradients.

% //////// TRAINING ----------------------------------------------------------------------------------

\subsection{Training}\label{sec:training}

As already suggested before, we usually employ \acp{RTN} in order to compute embeddings for
individuals that are used as input for some specific prediction task.
Therefore, it makes sense to train an \ac{RTN} together with the model that is used for computing
these predictions, and whenever we talk about an \ac{RTN} in the sequel, we shall assume that it is
used together with some predictor on top of it.
If we only care about individual embeddings irrespective of any particular subsequent task, then we
can simply add a feed-forward layer---or some other differentiable learning model---on top of the
\ac{RTN}, and train the model to reconstruct the provided feature vectors.
This way, an \ac{RTN} can be used as a kind of relational autoencoder.

Training such a model is straightforward, and switches back and forth between computing embeddings
and making predictions based on them.
In each training iteration, we start from the feature vectors of the individuals as they are provided
in the dataset.
Then, as a first step, we sample mini-batches of triples from the dataset, and randomly update the
current embedding of one of the individuals in each triple by means of our \ac{RTN}.
The total number of mini-batches that are considered in this step is a hyperparameter, and we found
during our experiments that it is in general not necessary to consider the entire dataset.

Next, we sample mini-batches of individuals from the dataset, and compute predictions for them based
on the embeddings that we created in the previous step.
In doing so, it makes sense to consider both individuals that have been updated as well as some that
still have their initial feature vectors as embeddings.
This is important for the model to learn how to deal with individuals that are involved in very few
relations or maybe no one at all, which is not a rare case in practice.
Therefore, in our experiments, we used mini-batches that were balanced with respect to this, and
switched back to step number one as soon as each of the previously updated individuals has been
sampled once.

The loss function as well as the optimization strategy employed depends, as usual, on the concrete
task, and is chosen case by case.

% //////// Related Models ----------------------------------------------------------------------------

\subsection{Related Models}

In the field of \ac{SRL}, there exist a few other approaches that model the effects of relations on individual embeddings in terms of
(higher-order) tensor products---cf., e.g., \citet{Nickel2011,Nickel2012}.
However, these methods, which belong to the category of latent variable models, are based on the idea of factorizing a tensor that
describes the structure of a relational dataset into a product of an embedding matrix as well as another tensor that represents the
relations present in the data.
The actual learning procedure is then cast as a regularized minimization problem based on this formulation.
In contrast to this, an \ac{RTN} computes embeddings, both during training and application, by means of a random process, and is thus
fundamentally different from this idea.

% //////// REASONING WITH RTNS ///////////////////////////////////////////////////////////////////////

\section{Reasoning with RTNs}\label{sec:reasoning}

% //////// APPLYING RTNS TO OKBS ---------------------------------------------------------------------

\subsection{Applying RTNs to OKBs}

As discussed in Section \ref{sec:okbs}, \acp{OKB} can be viewed as \acp{DAG}, and thus the
application of an \ac{RTN} to this kind of data is straightforward.
Therefore, we are only left with specifying the prediction model that we want to use on top of the
\ac{RTN}.
In the context of an \ac{OKB}, there are two kinds of predictions that we are interested in, namely
the membership of individuals to classes, on the one hand, and the existence of relations, on the
other hand.
From a \ac{ML} perspective, these are really two different targets, and we can describe them more
formally as follows: let $\mathcal{K}$ be an \ac{OKB}
that contains (exactly) the unary predicates $P_1 , \dotsc , P_k$ and (exactly) the binary
predicates $Q_1 , \dotsc , Q_\ell$, and $\mathcal{T} \subseteq \mathcal{K}$ the part of the \ac{OKB}
that we have as training set.
Then $t^{(1)}$ and $t^{(2)}$ are two target functions defined as
$$
    t^{(1)} :
    \left\{
    \begin{array}{l}
        individuals ( \mathcal{K} ) \to \{ -1 , 0 , 1 \}^k    \\
        i \mapsto \vect{x}^{( i )}
    \end{array}
    \right.
$$
and
$$
    t^{(2)} :
    \left\{
    \begin{array}{l}
        individuals ( \mathcal{K} )^2 \to \{ -1 , 0 , 1 \}^\ell    \\
        ( i , j ) \mapsto \vect{y}^{( i , j )}
    \end{array}
    \right.
$$
such that
$\vect{x}_m^{( i )}$ equals $1$, if $\mathcal{K} \models P_m ( i )$, $-1$, if
$\mathcal{K} \models \neg P_m ( i )$, and $0$, otherwise, and
$\vect{y}_m^{( i , j )}$ is defined accordingly with respect to $Q_m ( i , j )$.

Notice that all of the arguments of the functions $t^{(1)}$ and $t^{(2)}$ are individuals, and can
thus be represented as embeddings produced by an \ac{RTN}.
For computing actual predictions from these embeddings, we can basically employ an \ac{ML} model of
our choice.
In this work, however, we confine ourselves to multinomial logistic regression for $t^{(1)}$, i.e.,
we simply add a single feed-forward layer as well as a softmax on top it to the \ac{RTN}.
For $t^{(2)}$, we first add an additional original tensor layer as given in Equation
\ref{eq:original-tensor-layer}, like it was used by \citet{Socher2013}, and use multinomial
logistic regression on top of it as well.

% //////// PREDICTING CLASSES AND RELATIONS SIMULTANEOUSLY -------------------------------------------

\subsection{Predicting Classes and Relations Simultaneously}

While the targets $t^{(1)}$ and $t^{(2)}$ may be regarded as independent with respect to prediction,
this is clearly not the case for computing individual embeddings.
We require an embedding to reflect all of the information that we have about a single individual as
specified by the semantics of the considered \ac{OKB}.
Therefore, the tensor layers of an \ac{RTN} need to learn how to adjust individual vectors in view of
both unary and binary predicates, i.e., classes and relations.
To account for this, we train \acp{RTN}---facing the particular use case of ontology reasoning---on
mini-batches that consist of training samples for both of the prediction targets.

% //////// EVALUATION ////////////////////////////////////////////////////////////////////////////////

\section{Evaluation}\label{sec:evaluation}

To evaluate the suggested approach in a realistic scenario, we implemented a novel triple store, called \textbf{NeTS}
(\textbf{Ne}ural \textbf{T}riple \textbf{S}tore), that achieves ontology reasoning solely by means of an \ac{RTN}.
NeTS provides a simple, SPARQL-like, query interface that allows for submitting atomic queries as well as conjunctions of such (see 
Figure \ref{fig:query-example}).

\begin{figure}
    \selectfont
    \centering
    \ttfamily
    
    \begin{tabular}{l l}
        \multicolumn{2}{l}{NeTS> dbpedia:Person(?X),dbpedia:placeOfBirth(?X,?Y)}    \\
        \\
        ?X                          &?Y                                             \\
        =======================     &==============================                 \\
        dbpedia:Aristotle           &dbpedia:Stagira\_(ancient\_city)               \\
        dbpedia:Albert\_Einstein    &dbpedia:Ulm                                    \\
        $\vdots$                    &$\vdots$                                       \\
    \end{tabular}
    
    \caption{Example of a simple query in NeTS.}
    \label{fig:query-example}
\end{figure}

When the system is started, then the first step it performs is to load a set of learned weights from the disk---the actual learning process
is not part of NeTS right now, and may be incorporated in future versions.
Next, it observes whether there are previously generated embeddings of the individuals stored on disk already, and loads them as well, 
if any.
If this is not the case, however, then NeTS creates such embeddings as described above.
This step is comparable with what is usually referred to as materialization in the context of
database systems.
Traditionally, a database would compute all valid inferences that one may draw based on the provided
data, and store them somehow in memory or on  disk.
In contrast to this, NeTS accounts for these inferences simply by adjusting the individuals'
embeddings by means of a trained \ac{RTN}, which obviously has great advantages regarding its memory
requirements.
Note further that we do not store any actual inferences at this time, but rather compute them on
demand later on if this happens to become necessary.

Subsequent processing of queries is entirely based on these embeddings, and does not employ any kind
of formal reasoning at all.
This, in turn, allows for speeding up the necessary computations significantly, since we can dispatch
most of the the \MyQuote{heavy-lifting} to a GPU.

Our system is implemented in Python 3.4, and performs, as mentioned above, almost all numeric
computations on a GPU using PyCUDA 2016.1.2 \citep{Kloeckner2012}.
For learning the weights of our \acp{RTN}, we again used Python 3.4, along with TensorFlow 0.11.0
\citep{TensorFlow}.

% //////// TEST DATA ---------------------------------------------------------------------------------

\subsection{Test Data}

To maintain comparability, we evaluated our approach on the same datasets that \citet{Motik2014} used for their experiments with RDFox
\citep{Nenov2015}.\footnote{%
   All of these datasets are available at \url{http://www.cs.ox.ac.uk/isg/tools/RDFox/2014/AAAI/}.
}
As mentioned earlier, RDFox is indeed a great benchmark, since it has been shown to be the most efficient triple store at present.
For a comparison with other systems, however, we refer the interested reader to \citet{Motik2014}.

The test data consists of four Semantic Web \acp{KB} of different sizes and characteristics.
Among these are two real-world datasets, a fraction of DBpedia \citep{Bizer2009} and the Claros \ac{KB}\footnote{
    \url{http://www.clarosnet.org}
}, as well as two synthetic ones, LUBM \citep{Guo2005} and UOBM \citep{Ma2006}.
Their characteristics are summarized in Table \ref{tab:test-data}.

While all  these data are available in multiple formats, we made use of the ontologies specified in OWL and the facts provided as
n-triples for our experiments.
Furthermore, we considered only those predicates that appear for at least 5\% of the individuals in a database.
This is a necessary restriction to ensure that there is enough data for an \ac{RTN} to learn properly.
%\textcolor{red}{Notice, however, that this could be prevented by generating additional synthetic
%training facts, as mentioned in Section \ref{sec:reasoning}.}

\begin{table}[tb]
    \color{black}
    \centering
    \renewcommand\arraystretch{1.15}
	\begin{tabular}{l|*{4}{c|}}
	    \cline{2-5}
		                                                &\bf Claros    &\bf DBpedia    &\bf LUBM    &\bf UOBM    \\
		\hline
		\hline
		\multicolumn{1}{|l|}{\bf \ac{KRR} formalism}    &OWL           &OWL 2          &OWL         &OWL         \\
		\multicolumn{1}{|l|}{\bf \# of Individuals}     &6.5 M         &18.7 M         &32.9 M      &0.4 M       \\
		\multicolumn{1}{|l|}{\bf \# of Facts}           &18.8 M        &112.7 M        &133.6M      &2.2 M       \\
		\multicolumn{1}{|l|}{\bf \# of Classes}         &40 (13)       &349 (12)       &14 (4)      &39 (5)      \\
		\multicolumn{1}{|l|}{\bf \# of Relations}       &64 (20)       &13616 (16)     &13 (6)      &22 (11)     \\
		\hline
	\end{tabular}
	\vspace{.2cm}
	\caption{%
	    Characteristics of the test datasets.
	    All quantities refer to explicitly specified rather than inferred data, and the values in parentheses describe the classes and
	    relations, respectively, that appear with at least 5\% of the individuals.
	}
	\label{tab:test-data}
\end{table}

% //////// EXPERIMENTAL SETUP ------------------------------------------------------------------------

\subsection{Experimental Setup}

All our experiments were conducted on a server with 24 CPUs of type Intel Xeon E5-2620 (6$\times$2.40GHz), 64GB of RAM, and an Nvidia
GeForce GTX Titan X.
The test system hosted Ubuntu Server 14.04 LTS (64 Bit) with CUDA 8.0 and cuDNN 5.1 for GPGPU.
Notice, however, that NeTS does not make any use of multiprocessing or -threading besides GPGPU, which means that the only kind of
parallelization takes place on the GPU.
Therefore, in terms of CPU and RAM, NeTS had about half of the resources at its disposal that RDFox utilized in the experiments conducted by
\citet{Motik2014}.

Predicated on the use of the \ac{RTN} model, the datasets, including all of their inferences, were converted into directed graphs using
Apache Jena 2.13.0\footnote{
    \url{https://jena.apache.org}
}
and the OWL reasoner Pellet 2.4.0\footnote{
    \url{https://github.com/Complexible/pellet}
}---all of the import times reported in Table \ref{tab:comparison} refer to these graphs.
This reduced the size of the data, as stored on  disk, to approximately on third of the original dataset.
Furthermore, we removed a total of 50,000 individuals during training, together with all of the
predicates that these were involved in, as test set from each of the datasets, and similarly another
50,000 for validation---the results described in Table \ref{tab:ml-results} were retrieved for
these test sets.

% //////// RESULTS -----------------------------------------------------------------------------------

\subsection{Results}

In order to assess the quality of NeTS, we have to evaluate it on two accounts.
First, we need to consider its predictive performance based on the embeddings computed by the
underlying \ac{RTN} model, and second, we must ascertain the efficiency of the system with respect to
time consumption.

We start with the former.
To that end, consider Table \ref{tab:ml-results}, which reports the accuracies as well as F1 scores
that NeTS achieved on the held-out test sets, averaged over all classes and relations, respectively.
We see that the model consistently achieves great scores with respect to both measures.
Notice, however, that the F1 score is the more critical criterion, since all the predicates are
strongly imbalanced.
Nevertheless, the \ac{RTN} effectively learns embeddings that allow for discriminating positive from
negative instances.

\begin{table}[tb]
    \centering
    \renewcommand\arraystretch{1.15}
    \begin{tabular}{l | c c | c c |}
        \cline{2-5}
                                               &\multicolumn{2}{| c |}{\bfseries Classes}    &\multicolumn{2}{| c |}{\bfseries Relations}    \\
                                               &Avg.\ Accuracy    &Avg.\ F1                  &Avg.\ Accuracy    &Avg.\ F1                    \\
        \hline
        \hline
        \multicolumn{1}{| l |}{\bf Claros}     &0.969             &0.954                     &0.955             &0.942                       \\
        \multicolumn{1}{| l |}{\bf DBpedia}    &0.978             &0.959                     &0.961             &0.940                       \\
        \multicolumn{1}{| l |}{\bf LUBM}       &0.961             &0.948                     &0.959             &0.947                       \\
        \multicolumn{1}{| l |}{\bf OUBM}       &0.972             &0.953                     &0.973             &0.951                       \\
        \hline
    \end{tabular}
	\vspace{.2cm}
    \caption{The accuracies and F1 scores, averaged over all unary and binary predicates, respectively, for each dataset.}
    \label{tab:ml-results}
\end{table}

%In contrast to this, Table \ref{tab:comparison} lists the times that it took NeTS to import and materialize each of the datasets along with
Table \ref{tab:comparison}, in contrast, lists the times for NeTS to import and materialize each of the datasets along with
the respective measurements for RDFox \citep{Motik2014}.
As mentioned before, materialization refers to the actual computation of inferences, and usually depends on the expressivity of the ontology
as well as the number of facts available. %---the so-called data complexity.
We see that NeTS is significantly faster at the materialization step, while RDFox is faster at importing the data.
%Both conditions are explained easily.
This is explained as follows.
First, NeTS realizes reasoning by means of vector manipulations on a GPU, which is of course much faster than the symbolic computations
performed by RDFox.
As for the second point, RDFox makes use of extensive parallelization, also for importing data, while NeTS runs as a single process with
a single thread on a CPU.

However, from a practical point of view, materialization is usually more critical than import.
This is because an average database is updated with new facts quite frequently, while it is imported only once in a while.

Notice, however, that neither of the measures reported for NeTS contains the time for training the model.
The reason for this is that we train an \ac{RTN}, as mentioned earlier, with respect to an ontology rather than an entire \ac{OKB}.
Therefore, one can actually consider the training step as part of the setup of the database system.
For the datasets used in our experiments, training took between three and four days each.

\begin{table}[tb]
    \centering
    \renewcommand\arraystretch{1.15}
    \begin{tabular}{l | c c | c c c c |}
        \cline{2-7}
                                              &\multicolumn{2}{| c |}{\bfseries NeTS}    &\multicolumn{4}{| c |}{\bfseries RDFox}             \\
                                              &Import    &Materialization                &Import    &\multicolumn{3}{c |}{Materialization}    \\
        \hline
        \hline
        \multicolumn{1}{| l |}{\bf Claros}    &242       &28                             &48        &2062 &/    &---                          \\
        \multicolumn{1}{| l |}{\bf DBpedia}   &436       &69                             &274       &143  &/    &---                          \\
        \multicolumn{1}{| l |}{\bf LUBM}      &521       &52                             &332       &71   &/    &113                          \\
        \multicolumn{1}{| l |}{\bf OUBM}      &9         &11                             &5         &467  &/    &2501                         \\
        \hline
    \end{tabular}
	\vspace{.2cm}
    \caption{%
        The times for import and materialization (in seconds).
        For RDFox, these are the numbers reported by \citet{Motik2014} for computing a lower (left) and upper bound (right), respectively,
        on the possible inferences.
    }
    \label{tab:comparison}
\end{table}
% backup: 78, 99, 71, 51

% //////// SUMMARY & OUTLOOK /////////////////////////////////////////////////////////////////////////

\section{Summary and Outlook}

%\textcolor{red}{%
%    \begin{itemize}
%        \item In total, faster.
%        \item Much more memory-efficient.
%        \item Could be run on mobile GPUs.
%        \item Accuracy good, but possibly problematic in critical domains like medicine. (Although, could be even better with additionally
%            generated training data.)
%        \item Proof of concept!
%    \end{itemize}
%}

We have presented a novel method for \ac{SRL} based on deep learning, and used it to develop a highly
efficient, learning-based system for ontology reasoning.
Furthermore, we have provided an experimental comparison with one of the best logic-based ontology
reasoners at present, \mbox{RDFox}, on several large standard benchmarks, and showed that our approach
attains a high reasoning quality while being up to two orders of magnitude faster.

%An interesting topic for future research is to explore how the accuracy of ontology reasoning can be further 
%improved, e.g., by incorporating additional synthetic data and/or by a slightly refined \ac{RTN} architecture.
An interesting topic for future research is to explore ways to further improve our accuracy on
ontology reasoning.
This could be achieved, e.g., by incorporating additional synthetic data and/or slight refinements of
the \ac{RTN} architecture.
%A closely related topic is to further investigate the theoretical underpinning of the relational tensor network model.

%For future research, we discern two key challenges.%
%Firstly, investigating the application of the suggested approach in the context of real-world applications to identify issues outside our
%experimental setting.
%Secondly, further examination of the theoretical underpinning of the \ac{RTN} model.

% //////// ACKNOWLEDGMENTS ///////////////////////////////////////////////////////////////////////////

\subsubsection*{Acknowledgments}

This work was supported by the \ac{EPSRC}, under the grants EP/J008346/1, EP/L012138/1, and
EP/M025268/1, as well as the Alan Turing Institute, under the \ac{EPSRC} grant EP/N510129/1.
Furthermore, Patrick is supported by the \ac{EPSRC}, under grant OUCL/2016/PH, and the
Oxford-DeepMind Graduate Scholarship, under grant GAF1617\_OGSMF-DMCS\_1036172.

% //////// REFERENCES ////////////////////////////////////////////////////////////////////////////////
%
%\bibliographystyle{named}
%\bibliography{literature}

%% ================================================================================================ %%
%%  A C R O N Y M S                                                                                 %%
%% ================================================================================================ %%

\begin{acronym}
    \acro{BFS}{breadth-first search}
    \acro{BPTS}{backpropagation through structure}
    \acro{DAG}{directed acyclic graph}
    \acro{EPSRC}{Engineering and Physical Sciences Research Council}
    \acro{KB}{knowledge base}
    \acro{KG}{knowledge graph}
    \acro{KRR}{knowledge representation and reasoning}
    \acro{ML}{machine learning}
    \acro{NLP}{natural language processing}
	\acro{NN}{neural network}
	\acro{OKB}{ontological knowledge base}
	\acro{OWA}{open-world assumption}
	\acro{PBFS}{probabilistic breadth-first search}
	\acro{RNTN}{recursive neural tensor network}
	\acro{RSS}{random simple subgraph}
	\acro{RTN}{relational tensor network}
	\acro{SGD}{stochastic gradient descent}
	\acro{SRL}{statistical relational learning}
\end{acronym}

\end{document}